\documentclass{article}
\usepackage{spconf,amsmath,graphicx,hyperref}
\usepackage{amssymb,amsfonts}
\usepackage[nocompress]{cite}

\title{SetPrint: Black-Box LLM Fingerprinting via
Setwise Correspondence of Response Changes}
\name{%
\shortstack{
Jinmo Han \qquad Ju Yeon Kang \qquad Jimin Hong \\
Sunghwan Ahn \qquad Yonghyeon Jun \qquad Nam Soo Kim
}%
\thanks{This work was supported by a collaborative research project between LG Electronics, Inc. and Seoul National University (SNU).}}
\address{
Department of Electrical and Computer Engineering and INMC,\\
Seoul National University, Seoul, Republic of Korea \\
\texttt{\{jmhan, jykang, jmhong, shahn, yhjeon\}@hi.snu.ac.kr, nkim@snu.ac.kr}
}
\begin{document}

\maketitle
\begin{abstract}
Black-box LLM fingerprinting aims to determine whether a target model was derived from a source model when only the output responses to input queries are observable.
One way to reveal this relationship is to perturb each query and observe the resulting responses of both models, yielding two corresponding sets of response variations.
These sets provide two complementary sources of information: the relations among response changes within each set, and the correspondence of response changes between the two sets.
Existing approaches, however, have not fully exploited both.
We introduce \textbf{SetPrint}, a training-free method that jointly captures both within-set relations and between-set correspondence.
SetPrint expresses each response change relative to the mean change of its own set, and then compares changes from the two models that were induced by the same perturbation.
On LeaFBench, SetPrint consistently outperforms existing training-free black-box fingerprinting methods.
Further analyses attribute these gains to both within-set relations and between-set correspondence.
\end{abstract}

\begin{keywords}
LLM fingerprinting, black-box LLMs, model provenance, setwise correspondence, set comparison.
\end{keywords}

\section{Introduction}
\label{sec:intro}

Large language models (LLMs) are routinely fine-tuned~\cite{hu2022lora}, quantized~\cite{frantar2023gptq}, or otherwise adapted and then redistributed~\cite{nikolic2025mpt,shao2026zeroprint}.
Such redistribution makes it difficult to determine whether a deployed target model was derived from a given source model~\cite{shao2025sok,nikolic2025mpt}.
Identifying the target model's origin is important for tracing model provenance and auditing model reuse~\cite{shao2026zeroprint}.

\begin{figure}[t]
\centering
\includegraphics[width=\columnwidth]{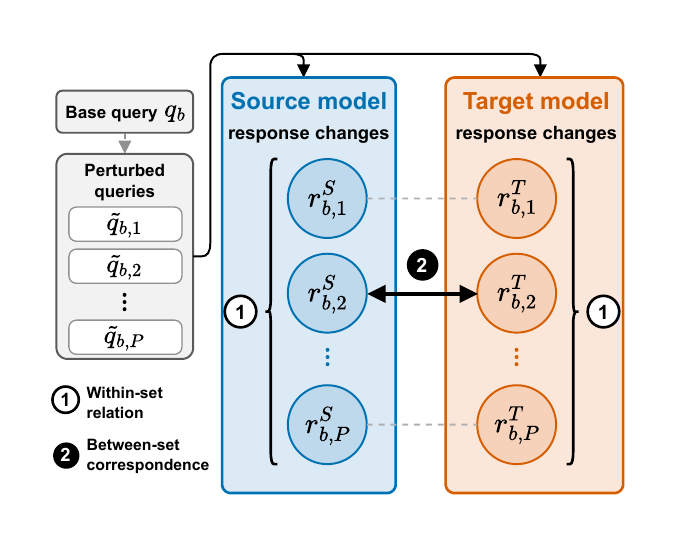}
\caption{Concept of SetPrint.
The same query perturbations yield two corresponding sets of response changes.
SetPrint uses both (1) within-set relations and (2) between-set correspondence, which pairs the changes induced by the same perturbation in the two models.}
\label{fig:fig1_concept}
\end{figure}

LLM fingerprinting identifies a model's origin from patterns in its responses or internal representations~\cite{shao2025sok,nikolic2025mpt}.
One family of methods embeds such a pattern into the source model so that it can later be recovered from a derived model~\cite{xu2024instructional,yamabe2025mergeprint}.
Such methods require the provider to modify the model in advance, and cannot be applied to models that have already been released.
A second family instead exploits patterns that an unmodified model already exhibits~\cite{shao2025sok,pasquini2025llmmap,gubri2024trap}.
It is further divided by how much access to the model is available: white-box methods use model parameters or hidden states~\cite{shao2025sok,zhang2025reef}, whereas black-box methods rely only on the responses returned for a query~\cite{pasquini2025llmmap,gao2025met,bai2025esf,ren2025cotsrf}.
This makes black-box methods applicable to real-world closed-API deployment, but identifying a model's origin from generated text alone remains challenging~\cite{shao2025sok,yang2026challenge}.

One of the simplest approaches in the black-box setting is to compare the responses that both source and target models produce for the same query set~\cite{shao2025sok,nikolic2025mpt,wu2026llmdna,yax2025phylolm}.
However, plain response similarity is not by itself reliable evidence, since independently developed models can still produce similar responses~\cite{nikolic2025mpt,yang2026challenge}.
One can instead obtain additional evidence by observing how a model's response changes when the query is modified~\cite{shao2026zeroprint,rauba2025statistical}.
Each modified version is given to the model, and its response is compared with the response to the original query~\cite{shao2026zeroprint,rauba2025statistical}.
Applying the same query modifications to the source and target models yields two sets of response changes, one for each model~\cite{shao2026zeroprint}.

There are two types of information in the two sets of response changes (Fig.~\ref{fig:fig1_concept}).
One is relations among changes within each set (within-set relations)~\cite{kriegeskorte2008representational,kornblith2019cka} and the other is pairings defined by the same modification across models (between-set correspondence)~\cite{wu2026tcf}.
Existing methods capture only part of this information.
One approach compares responses to each query separately, without accounting for within-set relations~\cite{wu2026tcf}.
Another compares summary representations of the two sets, without directly comparing changes paired by between-set correspondence~\cite{shao2026zeroprint}.
Neither fully exploits both types of information and can overlook cues to the target model's origin.

We propose \textbf{SetPrint}, a training-free black-box LLM fingerprinting method that jointly captures within-set relations and between-set correspondence.
SetPrint first perturbs the same queries to obtain two corresponding sets of response changes.
Within each set, every response change is expressed relative to the set mean, thereby capturing its relationship to the other responses.
Across the two sets, each relative change is compared with the one induced by the same perturbation in the other model.
SetPrint aggregates these comparisons into a single score for judging whether the target model was derived from the source.

On LeaFBench~\cite{shao2025sok}, SetPrint scores highest on AUROC, TPR@1\%FPR, and pAUC among the evaluated training-free black-box baselines.
Further analyses show that both within-set relations and between-set correspondence contribute to performance.
\section{Method}
\label{sec:method}

\subsection{Construction of Response Change Sets}
\label{sec:changes}

Let $S$ and $T$ denote the source and target models, respectively.
We input the same $B$ base queries $\{q_b\}_{b=1}^{B}$ and $P$ perturbed versions $\{\widetilde q_{b,p}\}_{p=1}^{P}$ of each to both models.
To reduce the effect of sampling randomness, we represent the response of model $m\in\{S,T\}$ to query $q$ by averaging $N$ sampled response embeddings:
\begin{equation}
\mu_m(q)
=\frac{1}{N}\sum_{\ell=1}^{N}
\varphi\!\left(y_m^{(\ell)}(q)\right),
\label{eq:mean_response}
\end{equation}
where $y_m^{(\ell)}(q)$ is the $\ell$-th sampled response and $\varphi$ is a fixed text encoder.
Using the same encoder for both models places their responses in a common embedding space.
We measure each response change by subtracting the mean response embedding for the base query from that for its perturbed version, and normalize the difference so that perturbations with large changes do not dominate the comparison:
\begin{equation}
\begin{aligned}
\delta_{b,p}^m
&=\mu_m(\widetilde q_{b,p})-\mu_m(q_b),\\
r_{b,p}^m
&=\frac{\delta_{b,p}^m}
{\max\!\left(\|\delta_{b,p}^m\|_2,\epsilon\right)},
\end{aligned}
\label{eq:response_change}
\end{equation}
where $\epsilon=10^{-8}$ prevents division by zero.
Enumerating the $K=BP$ pairs $(b,p)$ by a single index $i\in\{1,\dots,K\}$, we collect the response changes of each model into a set $\mathcal R_m=\{r_i^m\}_{i=1}^{K}$, where $r_i^S$ and $r_i^T$ arise from the same perturbed query.

\subsection{Setwise Correspondence}
\label{sec:correspondence}

SetPrint uses two kinds of information: within-set relations and between-set correspondence. For the former, it uses the average difference between one response change $r_i^m$ and all response changes $r_j^m$ in the same set:
\begin{equation}
\widetilde r_i^m
= \frac{1}{K}\sum_{j=1}^{K}(r_i^m-r_j^m)
= r_i^m-\bar r^m,
\label{eq:set_relative}
\end{equation}
where $\bar r^m=\frac{1}{K}\sum_{j=1}^{K}r_j^m$ is the set mean.
Subtracting $\bar r^m$ removes the component common to all response changes in the set, so the relative change $\widetilde r_i^m$ retains only the part that varies with the perturbation. 

For between-set correspondence, SetPrint takes the inner product between the corresponding relative changes $\widetilde r_i^S$ and $\widetilde r_i^T$, and averages over all $i$:
\begin{equation}
\begin{aligned}
\tau(\mathcal R_S,\mathcal R_T)
&=\frac{1}{K}\sum_{i=1}^{K}
\left\langle\widetilde r_i^S,\widetilde r_i^T\right\rangle\\
&=\frac{1}{K}\sum_{i=1}^{K}
\left(r_i^S-\bar r^S\right)^\top
\left(r_i^T-\bar r^T\right).
\end{aligned}
\label{eq:setprint}
\end{equation}
In this way, $\tau$ measures between-set correspondence while accounting for within-set relations.
A derived model inherits much of its source's behavior, so for a given perturbation, the two models should produce similar response changes. Subtracting each model's set mean removes what is common to all its responses, leaving the perturbation-specific part that the two models should share.
Thus, $\tau$ is larger for a target derived from the source than for an unrelated target, and SetPrint declares that $T$ is derived from $S$ if $\tau(\mathcal R_S,\mathcal R_T)$ exceeds a threshold $\gamma$.

Expanding Eq.~\eqref{eq:setprint} shows how SetPrint differs from scoring each perturbation separately:
\begin{equation}
\begin{aligned}
\tau(\mathcal R_S,\mathcal R_T)
&=\frac{1}{K}\sum_{i=1}^{K}
\left\langle r_i^S,r_i^T\right\rangle
-\left\langle\bar r^S,\bar r^T\right\rangle\\
&=\frac{1}{K}\sum_{i=1}^{K}
\left\langle r_i^S,r_i^T\right\rangle
-\frac{1}{K^2}\sum_{i,j=1}^{K}
\left\langle r_i^S,r_j^T\right\rangle.
\end{aligned}
\label{eq:setprint_expanded}
\end{equation}
The first term separately aggregates the score of each perturbation, so it measures between-set correspondence without considering within-set relations.
By including the second term, which is the inner product of the two set means, the within-set relations of Eq.~\eqref{eq:set_relative} are brought into the score.
This term averages the similarity over all $K^2$ pairs $(r_i^S,r_j^T)$, including the $K^2-K$ pairs with $i\neq j$ that do not correspond.
Thus, the score is how far the similarity between corresponding changes exceeds that between arbitrary pairs.
Sec.~\ref{sec:analysis} measures the contribution of each term of Eq.~\eqref{eq:setprint_expanded}.
\section{Experimental Setup}
\label{sec:experiments}

\subsection{Setup}
\label{sec:setup}

We evaluate on LeaFBench~\cite{shao2025sok}, using 13 source models and 149 target model instances.
The targets include variants obtained through fine-tuning, quantization, or changes to decoding settings, as well as unrelated models to the source model.

For each source-target pair, we send the same base and perturbed queries to both models.
We use HumanEval~\cite{chen2021humaneval} queries truncated to the first 20 words, following~\cite{shao2026zeroprint},
We use $B=4$ base queries and construct $P=4$ perturbations per base query by randomly swapping adjacent characters in five words.
For each base and perturbed query, we sample $N=10$ responses.
This requires $B(P+1)N=200$ responses per model and yields $K=BP=16$ normalized response changes.
We use temperature $0.7$, top-$p$ $0.9$, and top-$k$ $50$ unless otherwise specified by LeaFBench.
At most $512$ new tokens were generated per response.
We use all-mpnet-base-v2~\cite{reimers2019sentencebert,song2020mpnet} as the text encoder $\varphi$ in Eq.~\eqref{eq:mean_response}.

We report mean $\pm$ sample standard deviation over five runs for AUROC, TPR@1\%FPR (maximum empirical TPR at FPR $\leq 0.01$), and standardized pAUC over FPR $\in [0,0.05]$.
None of these metrics depends on a particular choice of $\gamma$.

\subsection{Baselines and Ablations}
\label{sec:compared_methods}
 
\textbf{Baselines.}
We compare SetPrint with fingerprinting methods spanning the white-box (REEF~\cite{zhang2025reef}), supervised black-box (LLMmap~\cite{pasquini2025llmmap}, LLM DNA with SVM classification~\cite{wu2026llmdna}), and unsupervised black-box (MET~\cite{gao2025met}, SEF~\cite{shao2025sok}, ZeroPrint~\cite{shao2026zeroprint}) settings.

\textbf{Alternative set comparison measures.}
To evaluate how the choice of set comparison measure affects fingerprinting performance, we replace the SetPrint score with alternative measures while keeping the response changes fixed.
MMD~\cite{gretton2012kernel} and Wasserstein distance~\cite{peyre2019computational} (with squared Euclidean distance) treat the two sets as unordered samples and thus ignore the between-set correspondence.
RSA~\cite{kriegeskorte2008representational} (with Pearson correlation between pairwise-distance matrices) and CKA~\cite{kornblith2019cka} (alignment of centered Gram matrices) use only the correspondence to align within-set relations entry by entry, never comparing corresponding changes directly.
Average Cosine is the mean cosine similarity between corresponding changes without mean subtraction, i.e., the first term of Eq.~\eqref{eq:setprint_expanded}.
For MMD and CKA, we report results with a linear kernel and with an RBF kernel whose bandwidth is the median pairwise distance among all response changes.

\textbf{Kernel variants of SetPrint.}
The inner product in Eq.~\eqref{eq:setprint_expanded} is one of the simplest choices of similarity between two response changes.
To check whether this simplicity misses anything a nonlinear similarity would capture, we replace the inner product in Eq.~\eqref{eq:setprint_expanded} with a kernel $k$, keeping the same two-term structure so that both within-set relations and between-set correspondence are retained:
\begin{equation}
\tau_k(\mathcal R_S,\mathcal R_T)
=\frac{1}{K}\sum_{i=1}^{K} k\!\left(r_i^S,r_i^T\right)
-\frac{1}{K^2}\sum_{i,j=1}^{K} k\!\left(r_i^S,r_j^T\right).
\label{eq:setprint_kernel}
\end{equation}
The inner product $k(x,y)=x^\top y$ recovers Eq.~\eqref{eq:setprint_expanded}.
SetPrint (Cubic), SetPrint (Exponential), and SetPrint (Arcsine) denote Eq.~\eqref{eq:setprint_kernel} with $k(x,y)=(x^\top y)^3$, $\exp(x^\top y)$, and $\arcsin(x^\top y)$, respectively.
\section{Results and Analysis}
\label{sec:results_analysis}

\subsection{Fingerprinting Performance}
\label{sec:results}

\begin{table*}[t]
\centering
\small
\caption{Fingerprinting results on LeaFBench (mean$\pm$std). Methods are grouped by required model access and supervision. $\dagger$: reported in~\cite{shao2026zeroprint}; $\ddagger$: our runs. Bold: best among black-box, unsupervised methods.}
\label{tab:main}
\setlength{\tabcolsep}{2.5pt}
\renewcommand{\arraystretch}{1.05}
\begin{tabular*}{\textwidth}{@{\extracolsep{\fill}}lccccc@{}}
\hline
\textbf{Method} & \textbf{Access} & \textbf{Supervised} &
\textbf{AUROC} $\uparrow$ & \textbf{TPR@1\%FPR} $\uparrow$ & \textbf{pAUC} $\uparrow$ \\
\hline
REEF$^{\dagger}$ & White box & No & $0.896 \pm 0.021$ & $0.634 \pm 0.021$ & $0.832 \pm 0.019$ \\
\hline
LLMmap$^{\dagger}$ & Black box & Yes & $0.632 \pm 0.009$ & $0.253 \pm 0.032$ & $0.635 \pm 0.009$ \\
LLMDNA + SVM$^{\ddagger}$ & Black box & Yes & $0.806 \pm 0.011$ & $0.292 \pm 0.066$ & $0.678 \pm 0.023$ \\
\hline
MET$^{\dagger}$ & Black box & No & $0.661 \pm 0.003$ & $0.329 \pm 0.004$ & $0.658 \pm 0.003$ \\
SEF$^{\dagger}$ & Black box & No & $0.581 \pm 0.008$ & $0.282 \pm 0.012$ & $0.646 \pm 0.004$ \\
ZeroPrint$^{\dagger}$ & Black box & No & $0.720 \pm 0.013$ & $0.366 \pm 0.009$ & $0.683 \pm 0.005$ \\
ZeroPrint$^{\ddagger}$ & Black box & No & $0.721 \pm 0.024$ & $0.320 \pm 0.023$ & $0.667 \pm 0.009$ \\
\textbf{SetPrint} & Black box & No & $\mathbf{0.776 \pm 0.008}$ & $\mathbf{0.403 \pm 0.002}$ & $\mathbf{0.704 \pm 0.004}$ \\
\hline
\end{tabular*}
\end{table*}

Table~\ref{tab:main} compares SetPrint with the baseline fingerprinting methods.
Among the black-box methods without supervision, SetPrint shows the highest AUROC, TPR@1\%FPR, and pAUC.

\begin{table}[t]
\centering
\small
\caption{Alternative set comparison measures on the same response change sets.
Bold: proposed method.
Last block: kernel variants of SetPrint.}
\label{tab:comparison}
\setlength{\tabcolsep}{3pt}
\renewcommand{\arraystretch}{1.05}
\begin{tabular*}{\columnwidth}{@{\extracolsep{\fill}}lcc@{}}
\hline
\textbf{Measure} & \textbf{AUROC} $\uparrow$ & \textbf{TPR@1\%FPR} $\uparrow$ \\
\hline
MMD (Linear) & $0.711 \pm 0.008$ & $0.323 \pm 0.010$ \\
MMD (RBF) & $0.713 \pm 0.007$ & $0.330 \pm 0.011$ \\
Wasserstein & $0.719 \pm 0.011$ & $0.331 \pm 0.011$ \\
CKA (Linear) & $0.620 \pm 0.027$ & $0.089 \pm 0.015$ \\
CKA (RBF) & $0.610 \pm 0.025$ & $0.104 \pm 0.019$ \\
RSA (Pearson) & $0.623 \pm 0.025$ & $0.095 \pm 0.014$ \\
Average Cosine & $0.755 \pm 0.010$ & $0.333 \pm 0.028$ \\
\textbf{SetPrint} & $\mathbf{0.776 \pm 0.008}$ & $\mathbf{0.403 \pm 0.002}$ \\
\hline
SetPrint (Cubic) & $0.758 \pm 0.010$ & $0.377 \pm 0.019$ \\
SetPrint (Exponential) & $0.776 \pm 0.008$ & $0.398 \pm 0.008$ \\
SetPrint (Arcsine) & $0.776 \pm 0.008$ & $0.404 \pm 0.005$ \\
\hline
\end{tabular*}
\end{table}

We next assess whether SetPrint's scoring rule offers an advantage over alternative measures on the same response change sets.
Table~\ref{tab:comparison} reports the results for the measures described in Sec.~\ref{sec:compared_methods}.
All measures use the same response change sets and differ in the scoring rule.
MMD and Wasserstein, which ignore the between-set correspondence, fall below SetPrint.
Average Cosine directly compares corresponding response changes and outperforms CKA and RSA, which compare within-set relations.
SetPrint further improves both AUROC and TPR@1\%FPR over Average Cosine.

Replacing the inner product with a nonlinear kernel (last block of Table~\ref{tab:comparison}) yields similar performance with the exponential and arcsine kernels and lower performance with the cubic kernel.
We therefore keep the simpler inner product.

\subsection{Analysis}
\label{sec:analysis}

To test the contributions of between-set correspondence and within-set relations, we alter one term of Eq.~\eqref{eq:setprint_expanded} while holding the other fixed.

\begin{table}[t]
\centering
\small
\caption{Contribution of between-set correspondence in SetPrint.
The target changes are left in their original order (None), shuffled among perturbations of each base query (Within-base), or shuffled across the full set (Full-set).}
\label{tab:permutation}
\setlength{\tabcolsep}{3pt}
\renewcommand{\arraystretch}{1.05}
\begin{tabular*}{\columnwidth}{@{\extracolsep{\fill}}lcc@{}}
\hline
\textbf{Permutation} & \textbf{AUROC} $\uparrow$ & \textbf{TPR@1\%FPR} $\uparrow$ \\
\hline
None (matched) & $0.776 \pm 0.008$ & $0.403 \pm 0.002$ \\
Within-base & $0.746 \pm 0.006$ & $0.335 \pm 0.013$ \\
Full-set & $0.494 \pm 0.004$ & $0.050 \pm 0.004$ \\
\hline
\end{tabular*}
\end{table}

\textbf{Contribution of between-set correspondence.}
We shuffle the target changes, replacing $r_i^T$ with $r_{\pi(i)}^T$ in Eq.~\eqref{eq:setprint}.
This leaves both response change sets and their means unchanged, so only the first term of Eq.~\eqref{eq:setprint_expanded} is affected.
Within-base permutation shuffles target changes among perturbations of the same base query, whereas full-set permutation shuffles them across all $K$ indices.
For each setting, we average AUROC and TPR@1\%FPR over 128 random permutations within each run.
As shown in Table~\ref{tab:permutation}, within-base permutation lowers AUROC, and full-set permutation reduces it to near chance.
These results show that preserving between-set correspondence for individual perturbations improves fingerprinting performance.

\begin{figure}[t]
\centering
\includegraphics[width=\columnwidth]{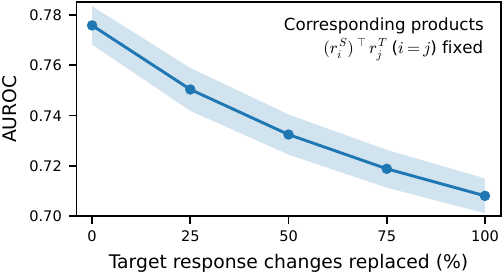}
\caption{Contribution of within-set relations in SetPrint.
In the products $(r_i^S)^\top r_j^T$ with $i\neq j$, target changes are progressively replaced by changes at the same indices from other models sharing the target's base model.
The products with $i=j$ remain fixed.
The line shows the mean AUROC across the five runs.
The shaded band shows sample standard deviation across these runs.}
\label{fig:fig2_setintervention}
\end{figure}

\textbf{Contribution of within-set relations.}
We test the contribution of within-set relations while keeping all products of corresponding changes $(r_i^S)^\top r_i^T$ fixed.
In the second term of Eq.~\eqref{eq:setprint_expanded}, we replace the target changes $r_j^T$ at randomly chosen indices $j$ only in the products $(r_i^S)^\top r_j^T$ with $i\neq j$.
For each model pair and each draw, we select one replacement model that shares the target's base model, excluding the source and target.
We use this model's change at each selected index $j$.
Within each run, we average AUROC over 48 random draws of replacement indices and models.
AUROC decreases as the replaced fraction grows (Fig.~\ref{fig:fig2_setintervention}), although every product of corresponding changes is preserved.
The within-set relations thus contribute to fingerprinting performance as well.
\section{Conclusion}
\label{sec:conclusion}

We introduced SetPrint, a training-free black-box LLM fingerprinting method that compares two models by how each responds to the same perturbed queries.
SetPrint centers each response change against the mean change of its own set, so that corresponding changes are compared in the context of the rest of the set.
This combines between-set correspondence with within-set relations in a single score.
On LeaFBench, SetPrint outperforms the evaluated training-free black-box baselines, and our analyses show that both kinds of information contribute.

\clearpage
\vfill\pagebreak
\ninept

\bibliographystyle{IEEEbib}
\bibliography{strings,refs_jy}

\end{document}